\documentclass{article}
\usepackage[nonatbib,preprint]{neurips_2019}
\usepackage[utf8]{inputenc} 
\usepackage[T1]{fontenc}    
\usepackage{hyperref}       
\usepackage{url}            
\usepackage{booktabs}       
\usepackage{amsfonts}       
\usepackage{nicefrac}       
\usepackage{microtype}      

\usepackage{microtype}
\usepackage{graphicx}
\usepackage{booktabs} 
\usepackage{amsmath}
\usepackage{amsfonts}
\usepackage{amsthm}
\usepackage{xcolor}

\usepackage{algorithm}
\usepackage{algorithmic}

\newtheorem{thm}{Theorem}

\title{Maximum Entropy Diverse Exploration: Disentangling Maximum Entropy Reinforcement Learning}

\author{Andrew Cohen \\ Binghamton University  \\acohen13@binghamton.edu \And 
    Lei Yu \\ Binghamton University \\ Yantai University \\ lyu@cs.binghamton.edu \And
    Xingye Qiao \\ Binghamton University \\ qiao@math.binghamton.edu \And 
    Xiangrong Tong \\ Yantai University \\ txr@ytu.edu.cn}

\begin{document}

\maketitle
\begin{abstract}
Two hitherto disconnected threads of research, diverse exploration (DE) and maximum entropy RL have addressed a wide range of problems facing reinforcement learning algorithms via ostensibly distinct mechanisms.  In this work, we identify a connection between these two approaches. First, a discriminator-based diversity objective is put forward and connected to commonly used divergence measures.  We then extend this objective to the maximum entropy framework and propose an algorithm Maximum Entropy Diverse Exploration (MEDE) which provides a principled method to learn diverse behaviors. A theoretical investigation shows that the set of policies learned by MEDE capture the same modalities as the optimal maximum entropy policy. In effect, the proposed algorithm disentangles the maximum entropy policy into its diverse, constituent policies. Experiments show that MEDE is superior to the state of the art in learning high performing and diverse policies.
\end{abstract}

\section{Introduction}
\label{introduction}
Deep reinforcement learning (RL) has demonstrated great potential by providing high performing control policies in a wide range of tasks from robotic manipulation to difficult games. However, the hard problem of exploration and issues therein such as data inefficiency, instability in the training process and susceptibility to local optima remain significant challenges. These problems are magnified in domains with sparse reward and multi-modal objective functions which are critical issues to overcome for real world scalability and implementation of RL solutions. 

Two hitherto disconnected threads of research to address the above challenges, diverse exploration (DE)~\cite{osband2016,Cohen-etal18,Cohen-etal19,DDDRL,Gang19} and maximum entropy RL~\cite{Fox-etal2015,A3C,DEBP,SAC} focus on alleviating these issues via ostensibly distinct mechanisms.  DE approaches generate diverse behavior policies that behave differently in similar states in order to achieve implicit exploration.  Numerous theoretical and empirical benefits in varied contexts of using diverse behavior policies have been demonstrated such as reducing the variance of on-policy gradient estimates~\cite{Cohen-etal19}. Alternately, maximum entropy RL augments the standard RL objective with a bonus for a highly stochastic behavior policy. Encouraging stochasticity promotes exploration and prevents early convergence. Likewise, maximum entropy policies provide various benefits such as composition for hierarchical tasks~\cite{HaarnICRA} and good starting policies for finetuning to more specific tasks.

Qualitatively interesting characteristics of maximum entropy policies have been noted~\cite{DEBP,SAC}: Policies trained with entropy bonuses tend to exhibit {\it all} reasonably good behaviors.  This is reminiscent of the motivation of DE approaches: there exist multiple, reasonably good but {\it different} behavior policies. Maximum entropy RL learns a single highly stochastic policy which implicitly captures different behavior modes whereas DE attempts to explicitly capture different behavior modes in distinct policies.  When viewed from this perspective, both approaches have a common end which suggests a potential connection.

On one hand, maximum entropy methods have an advantage in that they operate within a well defined theoretical framework which explains the resulting observed behavior whereas DE approaches typically utilize ad hoc or heuristic procedures such as bootstrapping data, adding diversity bonuses to the objective,  or directed network perturbations to generate diversity. On the other hand, DE methods have an advantage in that the diverse behaviors are distributed over separate policy approximations whereas maximum entropy methods provide diverse behaviors that are tangled together in a single stochastic policy.  Capturing different behaviors in separate policies could have significant benefit for problems like reward engineering and policy reuse in transfer learning as well as generally improving understandability of behaviors and the learning problem. Thus, a marriage of the two approaches may provide a principled method for capturing 
different behavior modes.

The contributions of this paper are 3-fold. {\it First}, by establishing a connection to KL divergence, it theoretically justifies a discriminator-based diversity objective that enables a computationally simple and effective method to encourage diversity among policies. {\it Second}, it extends this diversity objective to the maximum entropy framework and demonstrates formally that the optimal maximum entropy policy can be viewed as a mixture of policies that isolate distinct behavior modes. {\it Third}, it proposes a novel algorithm, Maximum Entropy Diverse Exploration (MEDE), to optimize the objective and capture these distinct behaviors in separate policies. The algorithm MEDE can be viewed as a disentangled version of maximum entropy reinforcement learning. 
\section{Preliminaries}
\label{prelim}
RL~\cite{sutton} problems are described by Markov Decision Processes (MDP)~\cite{puterman1994markov}. An MDP, $M$, is defined as a 5-tuple, $M =(S,A,P,\mathcal{R},\gamma)$, where $S$ is a fully observable set of states, $A$ is a set of possible actions, $P$ is the state transition model such that $P(s'|s,a) \in [0,1]$ describes the probability of transitioning to state $s'$ after taking action $a$ in state $s$, $\mathcal{R}_{s,s'}^a$ is the expected value of the immediate reward $r$ after taking $a$ in $s$, resulting in $s'$, and $\gamma\in (0,1)$ is the discount factor on future rewards. A {\it trajectory} of length $T$ is an ordered set of transitions: $\tau= \{s_0,a_0,r_1,s_1,a_1,r_2,...,s_{T-1},a_{T-1},r_{T}\}$.  
A solution to an MDP is a policy $\pi(a|s)$ which provides the probability of taking action $a$ in state $s$.
The performance of policy $\pi$ is the expected discounted return
\begin{align*}
J(\pi) = \mathbb{E}_{s_0,a_0..\sim \pi}[\sum_{t=0}^\infty \gamma^tr(a_t,s_t)]\ 
where\ s_0 \sim \rho(s_0),\  s_{t+1} \sim P(\cdot | s_t,a_t)
\end{align*}
and $\rho(s_0)$ is the distribution over start states.
The state-action value function and value function are defined as:
\begin{align*}
Q_{\pi}(s_t,a_t) = \mathbb{E}_{s_{t+1},a_{t+1}..}[\sum_{l=0}^\infty \gamma^lr(a_{t+l},s_{t+l})],\ 
V_{\pi}(s_t) = \mathbb{E}_{a_t,s_{t+1}..}[\sum_{l=0}^\infty \gamma^lr(a_{t+l},s_{t+l})]\\
\end{align*}

\subsection{Maximum entropy RL}
We present a general entropy regularized RL objective~\cite{equivalence}
\begin{equation}\label{ent_obj}
J(\pi)= \mathbb{E}_{s_0,a_0.. \sim \pi}[\sum_{t=0}^T r_t - \log(\pi(a_t| s_t)) + \log(\pi_0(a_t| s_t))].
\end{equation}
where $\pi_0$ is some prior or reference distribution over actions. Note, the entropy bonus in Eq.~\ref{ent_obj} is equal to the KL divergence $D_{KL}(\pi(\cdot | s_t)||\pi_0(\cdot | s_t))]$
but is presented as is for clarity of the coming theoretical analysis. Intuitively, this objective encourages the agent to maximize expected discounted return while also
trying to maintain an action distribution that is close to $\pi_0$.
The corresponding soft state-action and state value functions  are
\begin{align*}
Q_{soft}^{\pi}(s_t,a_t)= r_t + \gamma \mathbb{E}_{s_{t+1} \sim p(\cdot | s_t, a_t)}[V_{soft}^{\pi}(s_{t+1})]\\
V_{soft}^{\pi}(s_t)= \log \int_a \exp(Q_{soft}^{\pi}(s_t,a))\pi_0(a|s)da.\addtocounter{equation}{1}\tag{\theequation} \label{baseline_values}
\end{align*}
The optimal policy $\pi^*$ has the following form~\cite{ZiebartPHD}
\begin{equation}\label{ent_baseline_policy}
\pi^*(a | s) = \exp(Q_{soft}^{*}(s,a)-V_{soft}^{*}(s))\pi_0(a|s)
\end{equation}
where $Q_{soft}^{*}$ and $V_{soft}^{*}$ are the optimal soft state-action and state value functions. Note, $V_{soft}^{*}$ acts as the partition function normalizing the distribution. The probability that $\pi^*$ chooses a particular action is proportional to its exponentiated discounted expected return.
Under these definitions, the following is an equivalent form of $V_{soft}^{\pi}$
\begin{align}\label{eq_value}
V_{soft}^{\pi}(s_t)= \mathbb{E}_{a \sim \pi} [Q_{soft}^{\pi}(s_t,a) - \log(\pi(a| s_t)) + \log(\pi_0(a| s_t))].
\end{align}

Eq.~\ref{ent_obj} reduces to the objective more commonly studied in the literature~\cite{DEBP,SAC} up to a constant if $\pi_0$ is the uniform distribution. In what follows, we use $\pi^*_{soft}$ and $Q^*_{soft}$ to refer to the optimal policy and soft $Q$-function in the setting where $\pi_0$ is the uniform distribution and the agent is encouraged only to have a high entropy policy.

\section{Entropy and Diversity}
The goal of the maximum entropy framework is to learn a policy which assigns to each trajectory a probability proportional to the exponentiated sum of its rewards~\cite{levine}. Thus, when deploying the optimal $\pi^*_{soft}$, we observe a number of {\it diverse} behavior modalities.
In this section, we analyze a diversity objective within the theoretical framework of maximum entropy RL and demonstrate that it encourages a set of policies to each capture distinct modes of the trajectory distribution of $\pi^*_{soft}$ . Specifically, we introduce a surrogate for KL divergence, a common objective in DE approaches~\cite{Cohen-etal19, DDDRL}, to the maximum entropy objective to learn a set of distinct maximum entropy policies.  Then, we show that the optimal $Q^*_{soft}$ can be decomposed into the $Q$ function of any one of these individual policies plus a correction term based on the mixture of the individual policies. In effect, this shows how the multi-modality of $Q^*_{soft}$  is distributed over the $Q$ functions of a set of diverse policies.


\subsection{Discriminability as a Surrogate for Divergence}
To distinguish policies, we condition $\pi$ on discrete latent variables $z \sim p(z)$. Thus, each $z$ defines a distinct policy $\pi(a|s,z)$ which we denote by $\pi_z$. In this section, we relate the quantity $p(z|s,a)$, the {\it discriminability} of $z$ given a state-action pair $(s,a)$, to the pairwise KL divergence between policies conditioned on distinct $z$. When deploying policy $\pi_z$, if we can predict with high probability the latent variable $z$ from only the observed $(s,a)$, then the set of policies has high pairwise divergence. For intuition, the role of the discriminator in this context is analogous to that in the generative adversarial framework~\cite{GANS} where the distributions that must be distinguished are the action distributions with respect to different conditional variables. The key difference is that different behavior policies will try to {\it increase} their discriminability by maintaining action distributions with large divergence as per diversity objectives. This brings us to Theorem~\ref{kl} which considers the pairwise KL divergence between policies conditioned on only two distinct latent variables $z_i$ and $z_j$, however, the argument easily scales to an arbitrary number of policies. The proof is an application of Bayes' theorem and is contained in the supplement. 

\begin{thm}\label{kl}
Given a state $s$, increasing the discriminability of the variable $z$ increases the pairwise KL divergence 
$\hat{D}_{KL}(z_i, z_j) := D_{KL}(\pi(\cdot|s,z_i) || \pi(\cdot| s,z_j)) + D_{KL}(\pi(\cdot|s,z_j) || \pi(\cdot| s,z_i))$, $i\ne j$ i.e.
\begin{align*}
\hat{D}_{KL}(z_i, z_j)= 
\mathbb{E}_{a\sim \pi_{z_i}}[\log(\frac{p(z_i| s,a)}{p(z_j| s,a)})] + \mathbb{E}_{a\sim \pi_{z_j}}[\log(\frac{p(z_j| s,a)}{p(z_i| s,a)})].
\end{align*}
\end{thm}

DE approaches maximize divergence among policies typically by optimizing an explicit KL divergence objective. Theorem~\ref{kl} shows that this can be achieved by using a single, centralized discriminator.  As a diversity objective, the use of the discriminator is far less burdensome than computing divergence using explicit policy representations.  Similar discriminator-based methods have been used in past work without explicitly connecting to commonly used divergences~\cite{DIAYN,VIC,Haus18}.

\subsection{The MEDE Objective}
A degenerate way to become discriminable is for each policy to collapse to a different, deterministic policy. Thus, the maximum entropy setting is necessary in order for discriminability to be a meaningful diversity objective.  The objective is then
\begin{align}\label{z_obj}
J(\pi_z) =\mathbb{E}_{s_0,a_0.. \sim \pi_z}[\sum_{t=0}^T r_t - \log(\pi(a_t|s_t,z)) + \log(p(z | s_t,a_t))].
\end{align}
wherein each agent must trade-off entropy with discriminability.  The discriminator term is not a prior in the sense discussed in the Preliminaries since it is not a distribution over actions but the derivations of the optimal policy and value functions are unaffected.
The corresponding state value function and optimal policy become
\begin{align}\label{z_value}
V_{z}^{\pi}(s)= \log \int_a \exp(Q_{z}^{\pi}(s,a))p(z|s,a)da 
= \mathbb{E}_{a \sim \pi_z}[ Q_{z}^{\pi}(s,a) - \log(\pi(a| s,z)) + \log(p(z|s,a))]
\end{align}
\begin{align}\label{z_policy}
\pi^*(a | s,z) = \exp(Q_{z}^{*}(s,a) -V_{z}^{*}(s))p(z|s,a)
\end{align}
In what follows, we use $Q_{z}^{*}$ and $\pi^*_z$ to refer to the optimal soft $Q$ function and policy of Eqs.~\ref{z_value} and~\ref{z_policy}.

\subsection{Connecting DE and Entropy}
In our setting, a mixture policy arises naturally from marginalizing out $z$, $\pi_{/{\bf z}}(a|s) := \mathbb{E}_{z \sim p(\cdot|s)}[\pi(a|s,z)]$. $\pi_{/{\bf z}}$ is a multi-modal policy wherein each mode is represented by a distinct $\pi_z$ with relative proportions defined by the distribution $z \sim p(\cdot|s)$. To formally investigate the notion that $\pi^*_{soft}$ can be viewed as the mixture of diverse policies, we present two theorems using analysis techniques similar to those contained in~\cite{HaarnICRA} which characterize the structure of both $Q^*_{soft}$ and the optimal soft $Q$ function of $\pi_{/{\bf z}}$. We show that the multi-modality of both value functions can be decomposed into the modes captured by a single individual and a correction term which represents the modalities captured by the remaining individuals.  Finally, we show that the soft $Q$ function of the mixture $\pi_{/{\bf z}}$ is nearly equal to $Q^*_{soft}$ but for a KL divergence discrepancy that is actively minimized. 

We build up to the first result by discussing intuition regarding the mixture policy $\pi_{/{\bf z}}$.
Deploying $\pi_{/{\bf z}}$ entails, at each time step, sampling $z \sim p(\cdot|s)$ and then $a \sim \pi(\cdot|s,z)$. Then, given a state $s$, $\pi_{/{\bf z}}$ performs like policy $\pi_z$ with probability $p(z|s)$. At one extreme, if $p(z_i|s) = 1$ (implying $p(z_j|s)=0$, $j\ne i$) then $\pi_{/{\bf z}}$ performs exactly like the individual $\pi_{z_i}$. This would occur if state $s$ is {\it only} visited by $\pi_{z_i}$.  At the other, if $p(z_i|s) = p(z_j|s),\ \forall i,j$ then $\pi_{/{\bf z}}$ performs like each $\pi_z$ with equal probability. This occurs if state $s$ is an initial state or a bottleneck state which all policies must visit equally often.
This brings us to Theorem~\ref{soft_equal} which shows that $Q_{soft}^{*}$ is exactly the sum of the soft Q-function of an individual policy $Q^*_{z}$ and a correction term $S^*_z$ which represents the modes of the other individual policies as they are represented in the mixture $\pi_{/{\bf z}}$.
\begin{thm}\label{soft_equal}
\begin{align}\label{soft_equal_equation}
Q_{soft}^{*}(s,a) = Q^*_{z}(s,a) + S^*_z(s,a)
\end{align}
where $S^*_z(s,a)$ is the fixed point of the recursion 
\begin{align}\label{s_star}
S^{(k+1)}(s,a) = \gamma\mathbb{E}_{s' \sim p(s'|s,a)}[-\log(p(z|s'))+ \log\mathbb{E}_{a' \sim \pi_{/{\bf z}}(\cdot|s')}[\exp(S^{(k)}(s',a'))]]
\end{align}
\end{thm}
The recursive form of $S^*_z(s,a)$ indicates that it is the optimal soft Q-function of a policy $\pi_C(a|s)$  whose objective 
is to maximize the expected discounted sum of $-\log(p(z|s))$ while also minimizing its KL-divergence with the mixture policy $\pi_{/{\bf z}}$, $D_{KL}(\pi_C(\cdot|s)||\pi_{/{\bf z}}(\cdot|s))$ (see Eqs.~\ref{baseline_values} and~\ref{eq_value}). Then, $S^*_z(s,a)$ can be written in a more understandable form as 
\begin{align}\label{int_soft}
S^*_z(s,a) = \gamma\mathbb{E}_{s' \sim p(s'|s,a)}[-\log(p(z|s'))+ \mathbb{E}_{a' \sim \pi_C(\cdot|s')}[S^*_z(s',a') - \log(\pi_C(a'|s')) + \log(\pi_{/{\bf z}}(a'|s'))]].
\end{align}


Only the actions that are both high probability under $\pi_{/{\bf z}}$ and lead to states with low values of $p(z|s)$ (maximizes $-\log(p(z|s))$) have high value with respect to $S_z^*$. The actions $a \sim \pi^*_z$ may have high probability under $\pi_{/{\bf z}}$ but lead to states with high values of $p(z|s)$ and so the expected discounted sum of $-\log(p(z|s))$ will be close to zero.  Then, $\pi_C$ is like the mixture with $\pi^*_z$ removed and $S^*_z$ represents the modes captured by the other individual policies. $S^*_z$ can also be thought of as the "complement" to $Q^*_z$ in that it represents value for the modes of $\pi^*_{soft}$ that are not captured by $\pi^*_z$.
As a special case, $S^*_z\rightarrow 0$ as $p(z|s)\rightarrow 1$ because $\pi_{/{\bf z}}$ becomes dominated by $\pi^*_z$.  Thus, $\pi^*_z$ approaches $\pi^*_{soft}$ as it becomes more discriminable which we demonstrate in the experimental section. The two terms of the right hand side of Eq.~\ref{soft_equal_equation} in Theorem~\ref{soft_equal} show that individual policies isolate distinct modes of $\pi^*_{soft}$.
We now characterize the soft $Q$-function of the mixture $\pi_{/{\bf z}}$ to compare with $Q_{soft}^{*}$.
\begin{thm}\label{mix_equal}
Let $Q^*_{/{\bf z}}(s,a)$ be the optimal soft Q function of the mixture policy $\pi_{/{\bf z}}$. Then,
\begin{align*}
Q^*_{/{\bf z}}(s,a) = Q_{z}^{*}(s,a) + M^*_z(s,a) 
\end{align*}
where $M^*_z(s,a)$ is the fixed point of the recursion 
\begin{align}\label{m_star}
M^{(k+1)}(s,a) = \gamma\mathbb{E}_{s' \sim p(s'|s,a)}[-\log(p(z|s'))+ \mathbb{E}_{a' \sim \pi_{/{\bf z}}(\cdot|s')}[M^{(k)}(s',a'))]]
\end{align}
for any $z$.
\end{thm}
The discrepancy between $Q^*_{soft}$ and $Q^*_{/{\bf z}}$ is the softmax in Eq.~\ref{s_star} (or lack thereof in Eq.~\ref{m_star}).
From Eq.~\ref{int_soft}, $M^*_z(s,a) = S^*_z(s,a)$ when $\pi_{/{\bf z}}=\pi_C$. $\pi_C$ is trying to minimize the divergence $D_{KL}(\pi_C(\cdot|s)||\pi_{/{\bf z}}(\cdot|s))$ as part of its objective and so it is actively trying to minimize the discrepancy between $M^*_z$ and $S^*_z$. We leave further analysis of this discrepancy to future work.  Theorem~\ref{soft_equal} and~\ref{mix_equal} together show that $Q^*_{soft}$ is quite nearly the optimal $Q$ function of a mixture of diverse policies. Additionally, this also demonstrates that the mixture policy $\pi_{/{\bf z}}$ may serve as a reasonable approximation to the maximum entropy policy.

\section{Optimizing the Objective}

In practice, computing $p(z|s,a)$ would require integration over the entire state-action space which is generally intractable. So, we replace this term with a learned discriminator $q_{\rho}(z|s,a)$ parameterized by $\rho$ to obtain the variational lower bound~\cite{VB}:
\begin{align}\label{objf}
\hat{J}(\pi_z) = \mathbb{E}_{s_0,a_0.. \sim \pi_z}[\sum_{t=0}^T r_t - \log(\pi(a_t|s_t,z)) + \log(q_{\rho}(z | s_t,a_t))].
\end{align}
In order to condition the policy and value function networks on the variable $z$, we concatenate a one-hot encoding of $z$ to the state vector~\cite{DIAYN}.
We update the state-action value function network to minimize the soft Bellman residual as outlined in Soft-Actor-Critic (SAC)~\cite{SAC,HaarnApp} using the objective
\begin{align*}
J_Q(\theta) = \mathbb{E}_{(a_t,s_t,z) \sim \mathcal{D}}[\frac{1}{2} (Q_{\theta}(a_t,s_t,z) - (r_t + \mathbb{E}_{s_{t+1} \sim p(\cdot | s_t,a_t)}[V_{\theta}(s_{t+1},z)])^2]
\end{align*}
where $\mathcal{D}$ is the set of collected samples.
We do not learn an explicit network for $V$ but instead use the $Q$ networks to estimate $V$ according to Eq.~\ref{z_value} as in~\cite{HaarnApp}.
The discriminator optimizes the cross-entropy loss between its output vector and the one-hot encoding of $z$. As in~\cite{SAC},
we reparameterize the policy with a neural network transformation $a_t = f_{\phi}(\epsilon_t;s_t)$ to obtain the following objective
\begin{align*}
J_{\pi}(\phi) = \mathbb{E}_{(s_t,z)  \sim \mathcal{D}, \epsilon_t \sim \mathcal{N}}[\log(\pi_{\phi}(f_{\phi}(\epsilon_t;s_t)|s_t,z) - \log(q_{\rho}(z|s_t,f_{\phi}(\epsilon_t;s_t)))- Q_{\theta}(f_{\phi}(\epsilon_t;s_t),s_t,z))]
\end{align*}
with the corresponding gradient estimate
\begin{multline*}
\hat{\nabla}_{\phi}J(\phi) = \nabla_{\phi}\log(\pi_{\phi}(a_t|s_t,z)) + \\
(\nabla_{a_t}\log(\pi_{\phi}(a_t|s_t,z)) - \nabla_{a_t}\log(q_{\rho}(z|s_t,a_t))- \nabla_{a_t}Q_{\theta}(a_t,s_t,z))\nabla_{\phi}f_{\phi}(\epsilon_t;s_t)
\end{multline*}
where $a_t$ is evaluated at $f_{\phi}(\epsilon_t;s_t)$. Of particular interest is the term $\nabla_{a_t}\log(q_{\rho}(z|s_t,a_t))$ which indicates that this propagates a gradient through the discriminator network.  We found this to improve both the diversity of policies and the robustness to random seeds over alternate formulations where the discriminator did not contribute a gradient.

Eq.~\ref{objf} is very similar to the objective of Diversity is All You Need (DIAYN)~\cite{DIAYN}, a state of the art algorithm in the domain of unsupervised skill discovery.  DIAYN replaces the reward function with a learned discriminator term $\log(q_{\rho}(z|s))$ and optimizes the standard entropy objective $\mathbb{E}_{\pi_z}[\sum_{t=0}^T \log(q_{\rho}(z|s_t))- \log(\pi(a_t|s_t,z))]$ where the conditional variable $z$ defines distinct policies. Using this to augment the reward function $\hat{r}_t = r_t + \log(q_{\rho}(z|s_t))$ as suggested in~\cite{DIAYN}, DIAYN represents a competitive baseline against which to test the ability of MEDE to quickly learn diverse policies. 
The difference between the MEDE objective in Eq.~\ref{objf} and the DIAYN objective is the inclusion of the action $a$ when conditioning the discriminator. This may seem a trivial difference, however, including the action $a$ was critical to the theoretical analysis provided in the previous section because it enabled exploitation of Bayes' theorem.  Additionally, in the formulation of MEDE, we choose to include the discriminator in place of a prior as opposed to augmenting the reward.  This critical choice leads to the contribution of a gradient when optimizing MEDE's objective in Eq.~\ref{objf}. This additional direction from the discriminator gives MEDE an advantage in quickly and robustly finding diverse behaviors.

\section{Experimental Evaluation}

\setlength{\tabcolsep}{1pt}
\begin{figure}
    \begin{tabular}[t]{cccc}
        \includegraphics[width=0.25\textwidth]{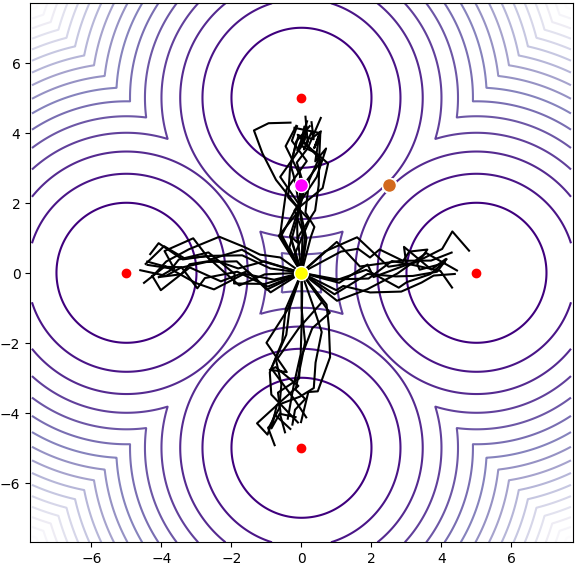} & \includegraphics[width=0.25\textwidth]{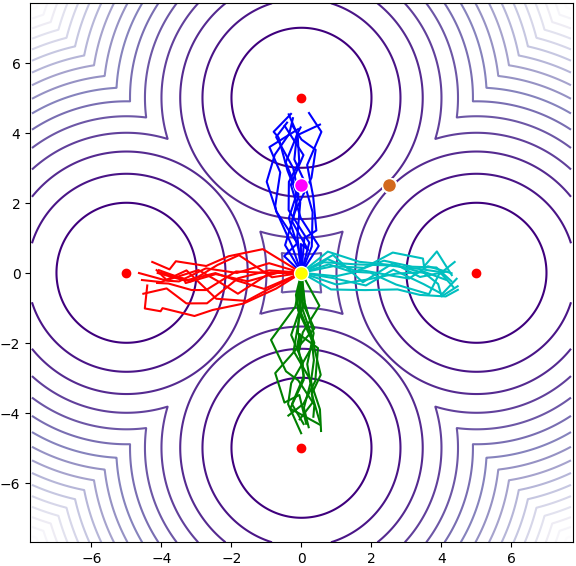}&
        \includegraphics[width=0.25\textwidth]{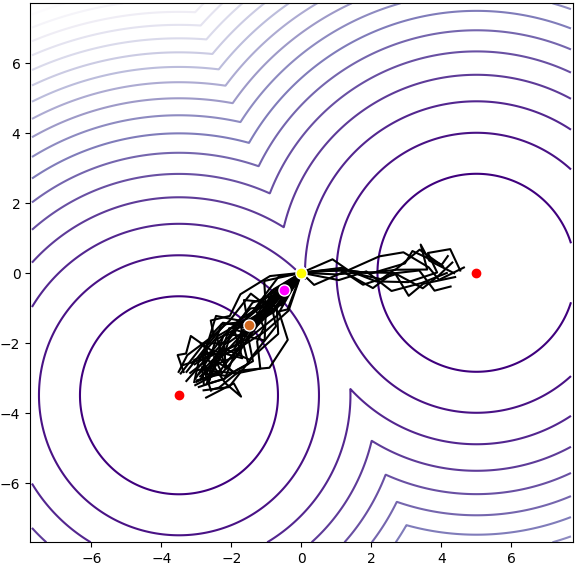} & \includegraphics[width=0.25\textwidth]{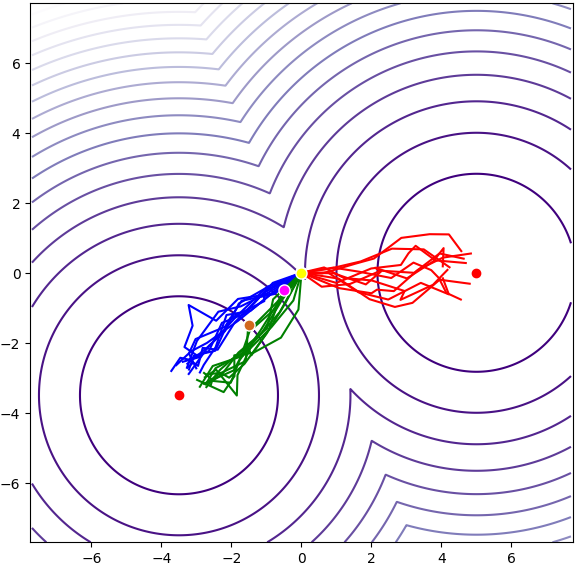}\\
        {\bf(a)} & {\bf(b)} & {\bf(c) } &{\bf(d) }\\ \midrule
    
    \multicolumn{4}{c}{
    \includegraphics[width=0.5\textwidth]{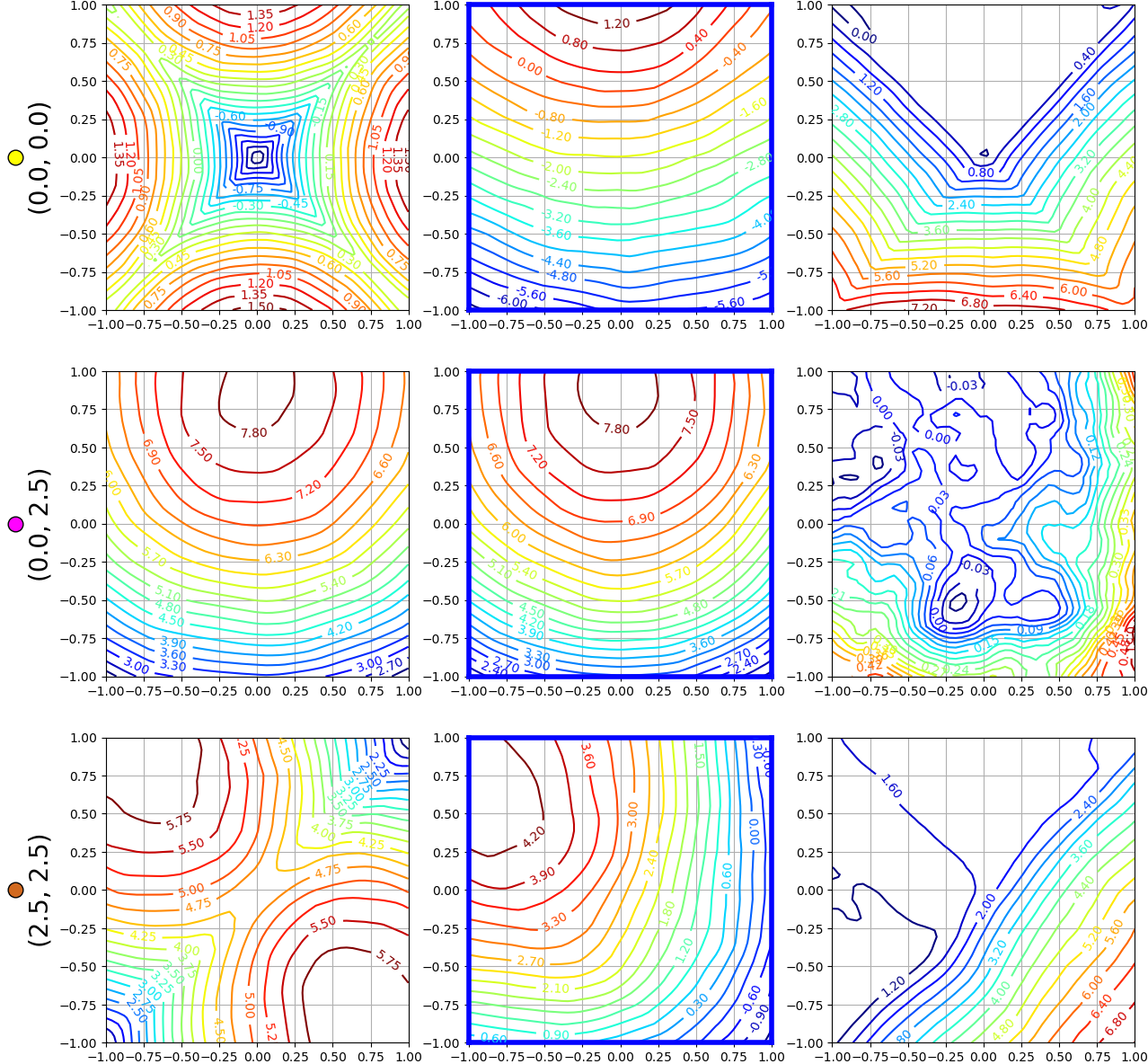}
    \includegraphics[width=0.5\textwidth]{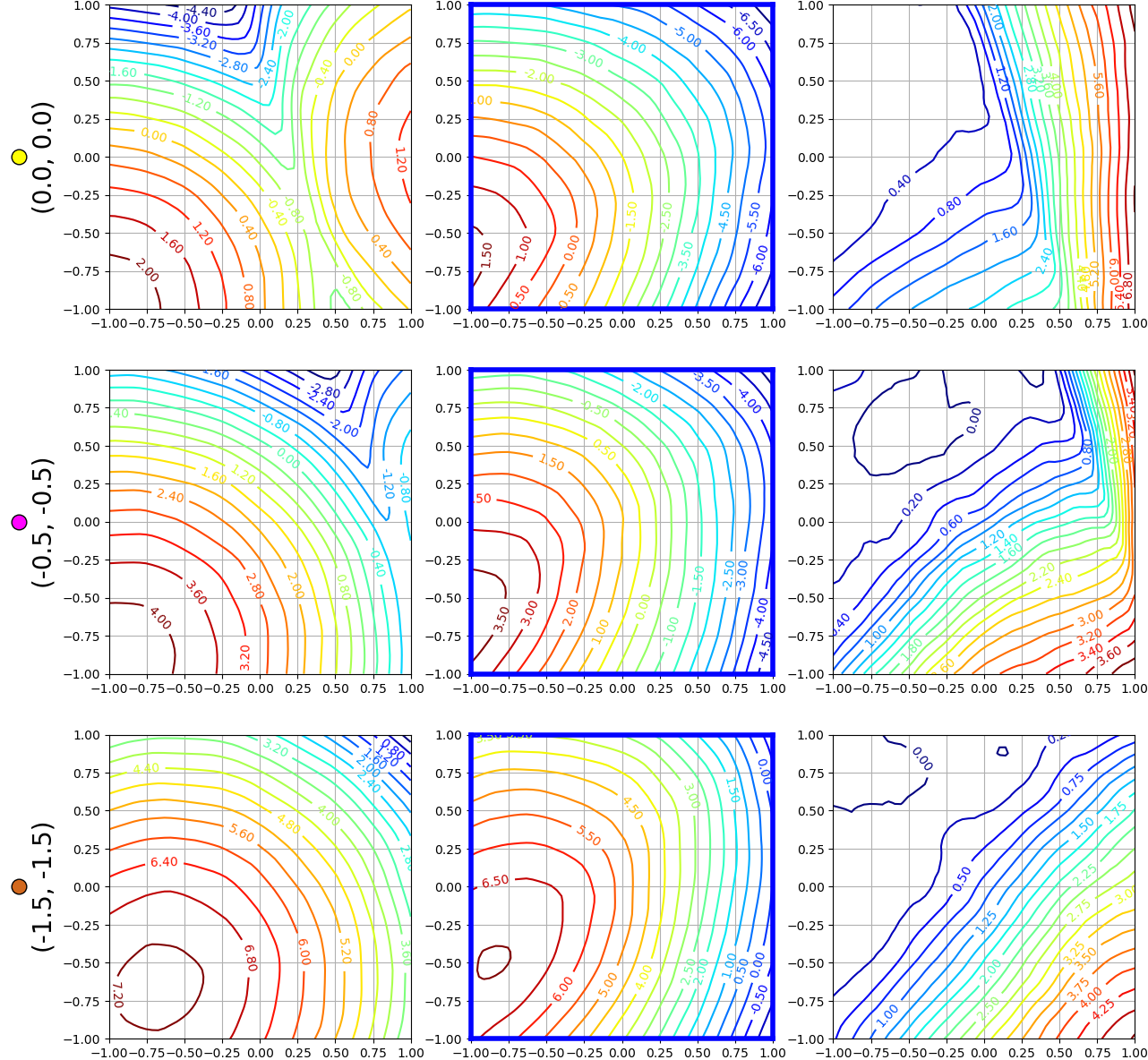}
}\\
    \multicolumn{2}{c}{{\bf (e)} $Q^*_{soft} - Q^*_{z} = S^*_z$} & \multicolumn{2}{c}{{\bf (f) } $Q^*_{soft} - Q^*_{z} = S^*_z$}
    \end{tabular}

        \caption{{\bf Above line:} SAC (a,c) and MEDE (b,d) on two instances of a multigoal environment. Goal states are marked by red dots. Colored lines represent the paths taken by the policy over rollouts. Different colors in figures (b,d) represent the policy conditioned on different values of $z$. {\bf Below line:} (e,f) Contours of the learned Q function for three different states marked by colored dots (yellow, pink, brown) in Figures (a-d) for SAC (left column), MEDE (middle, from the perspective of the 'blue' policy) and the difference (right). Each row corresponds to a different state. (e) corresponds to (a,b) and (f) corresponds to (c,d). $x$ and $y$ axes correspond to action dimensions.}
\label{multi}
\vspace{-3mm}
\end{figure}

In this experimental evaluation\footnote{Github url withheld to preserve anonymity}, we present results comparing MEDE to DIAYN and SAC on four different problems.  First, we present results in a simple multigoal environment which we use to provide a more intuitive understanding of the theory. Then, we present results on three difficult continuous control tasks; Hopper, Walker and Multidirection Ant~\cite{gym}. In these domains, we show that the performance of policies learned by MEDE and DIAYN are comparable to SAC and showcase the efficiency of MEDE in finding diversity. The experimental settings and hyperparameters are contained in the supplement. We denote the number of distinct policies with the cardinality $|Z|$. 
 
\subsection{Multigoal Environment}
We display the behavior of MEDE and SAC on two instances of a multigoal environment in which the objective is to navigate to within a threshold distance of any goal state (marked by red dots in (a,b,c,d) of Figure~\ref{multi}). The reward is the distance to the nearest goal state.  This objective has clearly identifiable modes (i.e. move to one of the goal states) that enables an investigation of the subtleties in the connection between MEDE and SAC.
In (a,b,c,d), colored lines represent paths taken by the policies learned by SAC and MEDE. The paths followed by the SAC policy (a,c) are colored black and the paths followed by the MEDE policies with different $z$ are differently colored (b,d). Beneath the top row, the 3 columns each of figures (e) and (f) correspond to the estimates of $Q^*_{soft}$, $Q^*_z$ and the difference $Q^*_{soft} - Q^*_z = S^*_z$ .  We present these three perspectives as an illustration of Theorem~\ref{soft_equal} to show how a mode of $Q^*_{soft}$ is isolated by $Q^*_z$ and the difference $S^*_z$ corresponds to the modes captured by the other policies.  We arbitrarily select the $Q^*_z$ of the 'blue' ($z=blue$) policy and denote this by highlighting the axes of the middle column in blue. Each row corresponds to a different state marked by a colored dot (yellow, pink and brown) in (a,b,c,d) and is also labeled with coordinates. 

 
Figures (a) and (b) show the behavior of SAC and MEDE with 4 equal goals and $|Z|=4$. To SAC, each goal is equally valuable but, to the blue policy, only the north goal has value because the value of $\log(p(blue|s,a))$ is extremely negative in the directions of the other three goals. The $Q$ estimates of the three states in (e) mirror this. In states that are equally close to different goals ((e), rows 1 and 3), $Q^*_{soft}$ is multimodal but $Q^*_z$ is unimodal and aimed at the north goal. The value contours of the difference $S^*_z$ correspond to the modes captured by the other policies.  In row 1 of (e), $S^*_z$ has three modes, of which the most valuable is in the direction of the south goal (most different from the blue policy). In row 3 of (e), $S^*_z$ is unimodal away from the north goal and towards the east goal because this corresponds to the only other policy with significant probability of visiting the state. Lastly, in row 2 of (e), only the blue policy has a significant probability of visiting the state so the value estimates are nearly identical with the difference $S^*_z$ being mostly approximation noise. This illustrates the notion discussed after Theorem~\ref{soft_equal} that $\pi^*_z$ approach $\pi^*_{soft}$ as they become discriminable.

Figures (c) and (d) display subtler aspects of the behavior of MEDE and SAC. In this instance, there are two goals where one (southwest goal) is slightly more valuable because it is closer to the initial state. We intentionally overprescribe the number of policies using $|Z|=3$ to illustrate MEDE's behavior in these circumstances. SAC visits the higher value goal more frequently and analogously two of the three MEDE policies learn to visit this mode. Additionally, the two policies become more deterministic (tighter grouping of paths) than the corresponding SAC paths and appear to split the mode. This can be seen quite clearly in the estimates of the $Q$ functions in figure (f) as the blue policy favors the top portion of the corresponding SAC mode. The progression of $S^*_z$ shows the diminishing effect of the red policy as it contributes less to the distribution of the mixture policy.  Experiments comparing MEDE to DIAYN on the first Multigoal domain are contained in the supplement. 

\vspace{-2mm}
\begin{figure}
    \begin{tabular}{cccc}
        \includegraphics[width=0.33\textwidth]{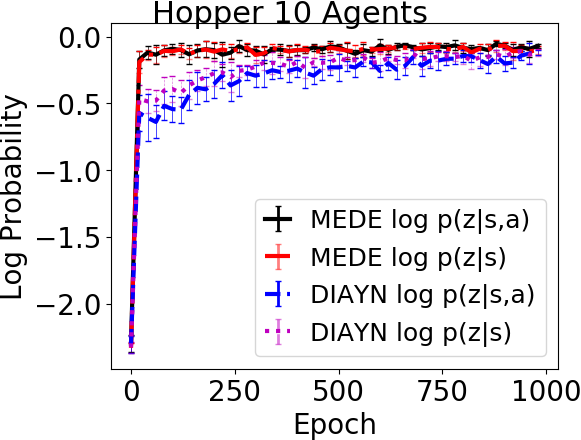} &
        \includegraphics[width=0.33\textwidth]{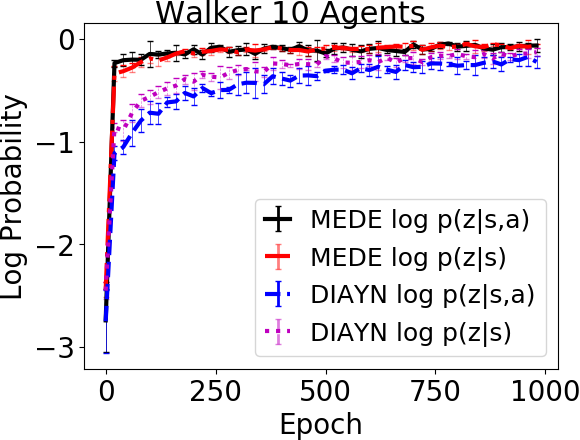} &
        \includegraphics[width=0.33\textwidth]{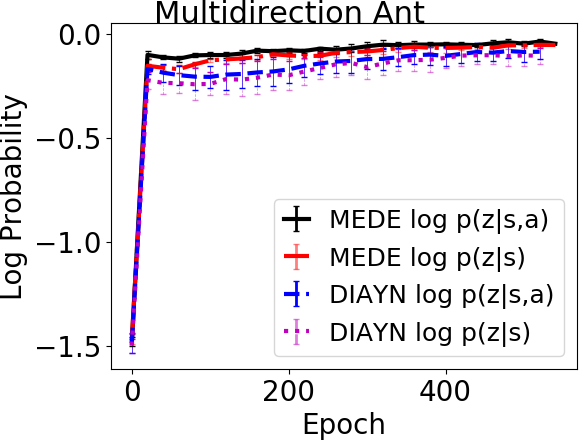} 
    \end{tabular}
\vspace{-2.5mm}
\caption{Average discriminability curves (both $\log p(z|s,a)$ and $\log p(z|s)$) and standard deviation of MEDE and DIAYN on Hopper, Walker and Multidirection Ant over 5 random seeds. In all three domains, MEDE more quickly finds discriminable policies and is more robust to random seed.}\label{disc_curves}
\vspace{-2.5mm}
\end{figure}

\begin{table}
\centering
\begin{tabular}{|c|c|c|c|c|c|} \hline
         & \multicolumn{2}{|c|}{Hopper} & \multicolumn{2}{|c|}{Walker}& M. Ant \\ \hline
         & Average & Max & Average & Max & Average\\ \hline
        MEDE 4 agents & $3155.6 \pm 116.2$ & $3197.5 \pm 107.4$ & $3728.7 \pm 231.8$ & $3778.8 \pm 237.6$ &$1996.2 \pm 224.9$  \\ \hline
        DIAYN 4 agents & $3120.0 \pm 80.0$ & $3170.0\pm 66.7$ & $3728.5 \pm 179.8$ & $3792.7 \pm 156.0$ &$1982.8 \pm 185.4$  \\ \hline
        MEDE 10 agents & $ 2657.0 \pm 140.0 $ & $2837.7 \pm 136.9 $ & $3283.8 \pm 194.9$ & $3420.3 \pm 165.3$ &-  \\ \hline
        DIAYN 10 agents & $2233.5 \pm 619.0$ & $ 2435.6 \pm 522.0$ & $3553.4 \pm 269.1$ & $3671.9 \pm 275.7$& - \\ \hline
        SAC & $3288.6 \pm 116.5$ & - & $3920.9 \pm 402.5$ & - & -  \\ \hline
\end{tabular}
\caption{Average and max return with standard deviation after learning of SAC, MEDE and DIAYN with 4 (Hopper, Walker and Ant) and 10 (Hopper and Walker) agents. Returns are comparable to SAC for 4 agents but it becomes increasingly difficult for MEDE and DIAYN to learn discriminable, good behaviors as the number of agents increases.}\label{returns}
\vspace{-2.5mm}
\end{table}

\subsection{Other Domains}
Figure~\ref{disc_curves} and Table~\ref{returns} aim to address the two points of investigation raised at the beginning of this section.  Our goal is to show that both MEDE and DIAYN can learn discriminable policies with performance comparable to that learned by SAC but that MEDE is superior to DIAYN in quickly and robustly learning diverse policies. Experimental results are provided for three difficult continuous control tasks; Hopper, Walker and Multidirection Ant which we refer to as M. Ant. M. Ant rewards speed in any direction in the the $xy$-plane and so significantly more diversity is available as compared to Hopper and Walker which only reward running in the positive x-direction. Then, finding diverse policies in Hopper and Walker is a greater challenge.

Figure~\ref{disc_curves} displays the average values of the discriminators $\log p(z|s,a)$ and $\log p(z|s)$ for both MEDE and DIAYN on the Hopper, Walker and M. Ant environment. The discriminability values are calculated over a minibatch sampled from the replay buffer at each iteration.  The sharp peak in the beginning of learning for all three curves corresponds to the discriminators being untrained at the beginning of learning. Note, the values of $\log p(z|s)$ and $\log p(z|s,a)$ are collected in MEDE and DIAYN, respectively, only for experimental purposes and are not used otherwise. In all three domains, MEDE immediately finds policies that are discriminable by both $(s,a)$ and $s$ whereas DIAYN only produces this behavior in M. Ant though still at a disadvantage to MEDE.  In Hopper and Walker, it takes nearly $500$ iterations for DIAYN to close the significance gap. We attribute this difference to the gradient from the discriminator which provides a highly efficient learning signal. We observe the same relative trends for 4 agents but we do not display figures due to space.

Table~\ref{returns} provides the average and max returns over $50$ trajectories per agent after learning.  The reported average is the average over all agents and the reported max is the average of the maximally performing agents from each experiment.  We do not report results for SAC on the M. Ant environment because the learned policy consistently converged to single direction across seeds. With 4 agents, both MEDE and DIAYN are able to learn policies that perform similarly to the SAC policy even with the disadvantage of trying to learn 4 distinct policies with equal samples and network parameters. As is clear from $10$ agents, it becomes increasingly difficult for both MEDE and DIAYN to learn policies as $|Z|$ increases and values over $20$ began to produce degenerate policies. We suspect this is a limitation of using a single network that toggles between policies by conditioning on latent variables (see Experimental details in the supplement). This is an issue we plan to address in future work. Figure~\ref{disc_curves} and Table~\ref{returns} demonstrate that both MEDE and DIAYN are capable of learning good policies but MEDE produces superior diversity. 

\section{Related Work}
Discovering diverse skills using discriminability objectives that differ from what is studied in this work has been applied to great effect in the domain of unsupervised skill discovery~\cite{DIAYN,VIC} and learning hierarchical policies to solve multiple tasks~\cite{Haus18}.  The novelty of our work is the formal connection of our objective to commonly used divergences and an analysis that illuminates some of the implicit dynamics of the standard maximum entropy formulation. 
Generating diverse behavior policies has been investigated as a means to enhance exploration. These approaches achieve diversity by optimizing an explicit KL divergence objective~\cite{Cohen-etal19,DDDRL}, bootstrapping trajectory data~\cite{Cohen-etal18} or applying Stein Variational Policy Gradient with different kernels~\cite{SVPG,Gang19}. 
Introducing further regularization to the maximum entropy framework has been investigated to achieve ends other than learning diverse behavior policies. \cite{Strouse18} use a mutual information regularizer in order to encourage/discourage an agent from sharing task-relevant information with another agent. \cite{Grau-Moya19} addresses exploration and performance of a single agent by learning a state-independent baseline that weights the relative importance of actions.

\section{Conclusion}
We have provided a novel theoretical investigation of Maximum Entropy RL which provides insight into the underlying mechanics of the framework. To that end, we developed and justified a novel diversity objective and proposed Maximum Entropy Diverse Exploration; a principled method that learns the behavior modes of the maximum entropy policy in separate policies. Lastly, we show experimentally how MEDE achieves this and also how MEDE more efficiently finds diverse policies that perform reasonably well than the existing state-of-the-art.
One future research direction is an investigation of the selection of $|Z|$ and whether this can be learned in a principled way. Another is to investigate policy network architectures that are optimized for representing diverse behaviors. Lastly, we will investigate principled methods of sharing trajectory data between agents.

\bibliography{Cohen_etal}

\begin{thebibliography}{10}

\bibitem{gym}
Greg Brockman, Vicki Cheung, Ludwig Pettersson, Jonas Schneider, John Schulman,
  Jie Tang, and Wojciech Zaremba.
\newblock Openai gym.
\newblock 2016.

\bibitem{Cohen-etal19}
A.~Cohen, X.~Qiao, L.~Yu, E.~Way, and X.~Tong.
\newblock Diverse exploration via conjugate policies for policy gradient
  methods.
\newblock In {\em Proceedings of the Thirty-Third AAAI Conference on Artificial
  Intelligence (AAAI'19)}, page to appear, 2019.

\bibitem{Cohen-etal18}
A.~Cohen, L.~Yu, and R.~Wright.
\newblock Diverse exploration for fast and reliable policy improvement.
\newblock In {\em Proceedings of the Thirty-Second AAAI Conference on
  Artificial Intelligence (AAAI'18)}, pages 2876--2883, 2018.

\bibitem{DIAYN}
B.~Eysenbach, A.~Gupta, J.~Ibarz, and S.~Levine.
\newblock Diversity is all you need: Learning diverse skills without a reward
  function.
\newblock In {\em International Conference on Learning Representations}, page
  to appear, 2019.

\bibitem{Fox-etal2015}
R.~Fox, A.~Pakman, and N.~Tishby.
\newblock Taming the noise in reinforcement learning via soft updates.
\newblock In {\em Uncertainty in Artificial Intelligence}, 2015.

\bibitem{Gang19}
T.~Gangwani, Q.~Liu, and J.~Peng.
\newblock Learning self-imitating diverse policies.
\newblock In {\em International Conference on Learning Representations}, page
  to appear, 2019.

\bibitem{GANS}
I.~Goodfellow, J.~Pouget-Abadie, M.~Mirza, B.~Xu, D.~Warde-Farley, S.~Ozair,
  A.~Courville, and Y.~Bengio.
\newblock Generative adversarial nets.
\newblock In {\em Proceedings of the 28th Conference on Neural Information
  Processing Systems}, 2014.

\bibitem{Grau-Moya19}
J.~Grau-Moya, F.~Liebfried, and P.~Vrancx.
\newblock Soft q-learning with mutual information regularization.
\newblock In {\em International Conference on Learning Representations}, page
  to appear, 2019.

\bibitem{VIC}
K.~Gregor, D.~Rezende, and D.~Wierstra.
\newblock Variational intrinsic control.
\newblock In {\em International Conference on Learning Representations}, 2016.

\bibitem{HaarnICRA}
T.~Haarnoja, V.~Pong, A.~Zhou, M.~Dalal, P.~Abbeel, and S.~Levine.
\newblock Composable deep reinforcement learning for robotic manipulation.
\newblock In {\em International Conference on Robotics and Automation}, 2018.

\bibitem{DEBP}
T.~Haarnoja, H.~Tang, P.~Abbeel, and S.~Levine.
\newblock Reinforcement learning with deep energy-based policies.
\newblock In {\em Proceedings of the 34th International Conference on Machine
  Learning}, 2017.

\bibitem{SAC}
T.~Haarnoja, A.~Zhou, P.~Abbeel, and S.~Levine.
\newblock Soft actor-critic: Off-policy maximum entropy deep reinforcement
  learning with a stochastic actor.
\newblock In {\em Proceedings of the 35th International Conference on Machine
  Learning}, 2018.

\bibitem{HaarnApp}
T.~Haarnoja, A.~Zhou, K.~Hartikainen, G.~Tucker, S.~Ha, J.~Tan, V.~Kumar,
  H.~Zhu, A.~Gupta, P.~Abbeel, and S.~Levine.
\newblock Soft actor-critic algorithms and applications.
\newblock 2018.

\bibitem{Haus18}
K.~Hausman, J.~Tobias Springenberg, Z.~Wang, N.~Heess, and M.~Riedmiller.
\newblock Learning an embedding space for transferable robot skills.
\newblock In {\em International Conference on Learning Representations}, 2018.

\bibitem{DDDRL}
Z.~Hong, A.~Shann, S.~Su, Y.~Chang, T.~Fu, and C.~Lee.
\newblock Diversity-driven exploration strategy for deep reinforcement
  learning.
\newblock In {\em Proceedings of the 32nd Conference on Neural Information
  Processing Systems}, 2018.

\bibitem{levine}
S.~Levine.
\newblock Reinforcement learning and control as probabilistic inference:
  Tutorial and review.
\newblock 2018.

\bibitem{SVPG}
Y.~Liu, P.~Ramachandran, Q.~Liu, and J.~Peng.
\newblock Stein variational policy gradient.
\newblock In {\em Uncertainty in Artificial Intelligence}, 2017.

\bibitem{A3C}
V.~Mnih, A.~P. Badia, M.~Mirza, A.~Graves, T.~Lillicrap, T.~Harley, D.~Silver,
  and K.~Kavukcuoglu.
\newblock Asynchronous methods for deep reinforcement learning.
\newblock In {\em Proceedings of the 33rd International Conference on Machine
  Learning}, 2016.

\bibitem{VB}
S.~Mohamed and D.~Rezende.
\newblock Variational information maximisation for intrinsically motivated
  reinforcement learning.
\newblock In {\em Proceedings of the 29th Conference on Neural Information
  Processing Systems}, pages 2125--2133, 2015.

\bibitem{osband2016}
I.~Osband, C.~Blundell, A.~Pritzel, and B.~Van Roy.
\newblock Deep exploration via bootstrapped dqn.
\newblock In {\em Proceedings of the 30th Conference on Neural Information
  Processing Systems}, 2016.

\bibitem{puterman1994markov}
M.L. Puterman.
\newblock {\em {Markov decision processes: Discrete stochastic dynamic
  programming}}.
\newblock John Wiley \& Sons Inc. New York NY USA, 1994.

\bibitem{equivalence}
J.~Schulman, X.~Chen, and P.~Abbeel.
\newblock Equivalence between policy gradents and soft q learning.
\newblock 2018.

\bibitem{Strouse18}
D.~Strouse, M.~Kleiman-Weiner, J.~Tenenbaum, M.~Botvinick, and D.~Schwab.
\newblock Learning to share and hide intentions using information
  regularization.
\newblock In {\em Proceedings of the 32nd Conference on Neural Information
  Processing Systems}, 2018.

\bibitem{sutton}
R.~S. Sutton and A.~G. Barto.
\newblock {\em Reinforcement Learning: An Introduction}.
\newblock {The MIT Press}, 1998.

\bibitem{ZiebartPHD}
B.~D. Ziebart.
\newblock Modeling purposeful adaptive behavior with the principle of maximum
  causal entropy.
\newblock In {\em PhD Thesis}, 2010.

\end{thebibliography}
\bibliographystyle{plain}
\pagebreak
\appendix

\setcounter{thm}{0}
\section{Theorem Proofs}
Note, by applying Bayes' $\frac{\pi(a | s,z)}{p(z|s,a)}= \frac{\pi_{/{\bf z}}(a | s)}{p(z|s)}$ to the optimal policy
\begin{align*}
\pi^*(a | s,z) = \exp(Q_{z}^{*}(s,a) -V_{z}^{*}(s))p(z|s,a)
\end{align*}
we obtain the relationship 
\begin{align*}
\pi_{/{\bf z}}(a | s) = \exp(Q_{z}^{*}(s,a) -V_{z}^{*}(s))p(z|s)\ (*)
\end{align*}
which we refer to with $(*)$ in the proofs of Theorems 2 and 3.

\subsection{Proof of Theorem 1}
\begin{thm}
Given a state $s$, increasing the discriminability of the variable $z$ increases the pairwise KL divergence 
$\hat{D}_{KL}(z_i, z_j) := D_{KL}(\pi(\cdot|s,z_i) || \pi(\cdot| s,z_j)) + D_{KL}(\pi(\cdot|s,z_j) || \pi(\cdot| s,z_i))$, $i\ne j$ i.e.
\begin{align*}
\hat{D}_{KL}(z_i, z_j)= 
\mathbb{E}_{a\sim \pi_{z_i}}[\log(\frac{p(z_i| s,a)}{p(z_j| s,a)})] + \mathbb{E}_{a\sim \pi_{z_j}}[\log(\frac{p(z_j| s,a)}{p(z_i| s,a)})].
\end{align*}
\begin{proof}
\begin{align*}
D_{KL}(\pi(\cdot|s,z_i) || \pi(\cdot| s,z_j)) = \mathbb{E}_{a\sim \pi_{z_i}}[\log(\frac{\pi(a| s,z_i)}{\pi(a| s,z_j)})].
\end{align*}
Note the following conditional and joint probability identities $p(a,z|s)=p(a|s,z)p(z|s)$ and $ p(a,z|s)= p(z|s,a)p(a|s)$.
Then, by substitution
\begin{align*}
\mathbb{E}_{a\sim \pi_{z_i}}[\log(\frac{\pi(a| s,z_i)}{\pi(a| s,z_j)})] = \mathbb{E}_{a\sim \pi_{z_i}}[\log(\frac{p(a,z_i| s) p(z_j|s)}{p(a,z_j| s)p(z_i|s)})]\\
=\mathbb{E}_{a\sim \pi_{z_i}}[\log(\frac{p(a,z_i| s)}{p(a,z_j| s)}) + \log(\frac{p(z_j|s)}{p(z_i|s)})]=\\
\mathbb{E}_{a\sim \pi_{z_i}}[\log(\frac{p(a,z_i| s)}{p(a,z_j| s)})] + \log(\frac{p(z_j|s)}{p(z_i|s)})
\end{align*}
where the second line follows from substituting the first identity and splitting the logarithm and the third line follows because $p(z_i|s)$ and $p(z_j|s)$ are independent of the action $a$. By substituting the second identity,
\begin{align*}
\mathbb{E}_{a\sim \pi_{z_i}}[\log(\frac{p(a,z_i| s)}{p(a,z_j| s)})] + \log(\frac{p(z_j|s)}{p(z_i|s)})=\\
\mathbb{E}_{a\sim \pi_{z_i}}[\log(\frac{p(z_i| s,a)p(a|s)}{p(z_j| s,a)p(a|s)})] + \log(\frac{p(z_j|s)}{p(z_i|s)})=\\
\mathbb{E}_{a\sim \pi_{z_i}}[\log(\frac{p(z_i| s,a)}{p(z_j| s,a)})] + \log(\frac{p(z_j|s)}{p(z_i|s)}).
\end{align*}
Likewise, the reverse KL divergence $D_{KL}(\pi(\cdot|s,z_j) || \pi(\cdot| s,z_i))$ is
\begin{align*}
\mathbb{E}_{a\sim \pi_{z_j}}[\log(\frac{p(z_j| s,a)}{p(z_i| s,a)})] + \log(\frac{p(z_i|s)}{p(z_j|s)}).
\end{align*}
The pairwise KL divergence is then
\begin{align*}
D_{KL}(\pi(\cdot|s,z_i) || \pi(\cdot| s,z_j)) + D_{KL}(\pi(\cdot|s,z_j) || \pi(\cdot| s,z_i))=\\
\mathbb{E}_{a\sim \pi_{z_i}}[\log(\frac{p(z_i| s,a)}{p(z_j| s,a)})] + \mathbb{E}_{a\sim \pi_{z_j}}[\log(\frac{p(z_j| s,a)}{p(z_i| s,a)})] 
\end{align*}
where the second terms cancel because $\log(\frac{p(z_i|s)}{p(z_j|s)}) = \log(p(z_i|s)) - \log(p(z_j|s))$.
\end{proof}
\end{thm}

\subsection{Proof of Theorem 2}
\begin{thm}
\begin{align*}
Q_{soft}^{*}(s,a) = Q^*_{z}(s,a) + S^*_z(s,a)
\end{align*}
where $S^*_z(s,a)$ is the fixed point of the recursion 
\begin{align*}
S^{(k+1)}(s,a) = \gamma\mathbb{E}_{s' \sim p(s'|s,a)}[-\log(p(z|s'))+ \log\mathbb{E}_{a' \sim \pi_{/{\bf z}}(\cdot|s')}[\exp(S^{(k)}(s',a'))]]
\end{align*}
\begin{proof}
Define $Q^{(0)}(s,a)=Q^*_{z}(s,a)$ and $S^{(0)}(s,a) = 0$ and assume  $Q^{(k)} = Q_{z}^{*}(s,a)+S^{(k)}(s,a)$ for some $k$. The base case (when $k=0$) is clearly satisfied. Then, applying the soft bellman backup 
\begin{align*}
Q^{(k+1)}(s,a) = r + \gamma \mathbb{E}_{s' \sim p(s'|s,a)}[\log \int_{a'} \exp(Q^{k}(s',a'))da']\\
= r + \gamma \mathbb{E}_{s' \sim p(s'|s,a)}[\log \int_{a'} \exp(Q^*_{z}(s',a')+S^{(k)}(s',a'))da']\\
= r + \gamma \mathbb{E}_{s' \sim p(s'|s,a)}[\log \int_{a'} \exp(Q^*_{z}(s',a'))\exp(S^{(k)}(s',a'))da']\\
= r + \gamma \mathbb{E}_{s' \sim p(s'|s,a)}[\log \int_{a'} \exp(V^*_{z}(s'))\frac{\pi_{/{\bf z}}(a'|s')}{p(z|s')}\exp(S^{(k)}(s',a'))da']\\
= r + \gamma \mathbb{E}_{s' \sim p(s'|s,a)}[V^*_{z}(s') - \log(p(z|s') + \log \int_{a'} \pi_{/{\bf z}}(a'|s')\exp(S^{(k)}(s',a'))da']\\
= Q^*_{z}(s,a) + S^{(k+1)}(s,a)
\end{align*}
where the fourth line follows from substiting with $(*)$, $\exp(V^*_{z}(s))\frac{\pi_{/{\bf z}}(a|s)}{p(z|s)}=\exp(Q_{z}^{*}(s,a))$.
Since the bellman update converges from any bounded $Q^{(0)}$, taking $k \to \infty$, we obtain
\begin{align*}
Q^*_{soft}(s,a) = Q^*_{z}(s,a) + S^*_z(s,a) 
\end{align*}
where $S^*_z(s,a)$ is the fixed point of the recursion 
\begin{align*}
S^{(k+1)}(s,a) = \mathbb{E}_{s' \sim p(s'|s,a)}[-\log(p(z|s'))+ \log\mathbb{E}_{a' \sim \pi_{/{\bf z}}(\cdot|s')}[\exp(S^{(k)}(s',a'))]
\end{align*}

\end{proof}
\end{thm}

\subsection{Proof of Theorem 3}
\begin{thm}
Let $Q^*_{/{\bf z}}(s,a)$ be the optimal soft Q function of the mixture policy $\pi_{/{\bf z}}$. Then,
\begin{align*}
Q^*_{/{\bf z}}(s,a) = Q_{z}^{*}(s,a) + M^*_z(s,a) 
\end{align*}
where $M^*_z(s,a)$ is the fixed point of the recursion 
\begin{align*}
M^{(k+1)}(s,a) = \gamma\mathbb{E}_{s' \sim p(s'|s,a)}[-\log(p(z|s'))+ \mathbb{E}_{a' \sim \pi_{/{\bf z}}(\cdot|s')}[M^{(k)}(s',a'))]]
\end{align*}
for any $z$.
\begin{proof}
Define $Q^{(0)}=Q_{z}^{*}(s,a)$ and $M^{(0)}(s,a) = 0$ and assume  $Q^{(k)} = Q_{z}^{*}(s,a)+M^{(k)}(s,a)$ for some $k$. The base case when $k=0$ is clearly satisfied. Then, applying the soft bellman backup 
\begin{multline}
Q^{(k+1)}(s,a) = r + \gamma \mathbb{E}_{s' \sim p(s'|s,a)}[\mathbb{E}_{a'\sim \pi_{/{\bf z}}} [Q^{k}(s',a') - \log(\pi_{/{\bf z}}(a'|s'))]]\\
Q^{(k+1)}(s,a) = r + \gamma \mathbb{E}_{s' \sim p(s'|s,a)}[\mathbb{E}_{a'\sim \pi_{/{\bf z}}} [Q^{k}(s',a') - Q_{z}^{*}(s',a') - \log(p(z|s')) + V_{z}^{*}(s')]]\\
=r + \gamma \mathbb{E}_{s' \sim p(s'|s,a)}[\mathbb{E}_{a'\sim \pi_{/{\bf z}}} [Q_{z}^{*}(s',a') + M^{(k)}(s',a') - Q_{z}^{*}(s',a') - \log(p(z|s')) + V_{z}^{*}(s')]]\\
=r + \gamma \mathbb{E}_{s' \sim p(s'|s,a)}[\mathbb{E}_{a'\sim \pi_{/{\bf z}}} [- \log(p(z|s')) + M^{(k)}(s',a')  + V_{z}^{*}(s')]\\
=Q_{z}^{*}(s,a) + \gamma \mathbb{E}_{s' \sim p(s'|s,a)}[- \log(p(z|s')) + \mathbb{E}_{a'\sim \pi_{/{\bf z}}} [M^{(k)}(s',a') ]]\\
=Q_{z}^{*}(s,a) + M^{(k+1)}(s,a)
\end{multline}
where the second line follows from substiting with $(*)$, $\pi_{/{\bf z}}(a|s)=\frac{\exp(Q_{z}^{*}(s,a))}{\exp(V_{z}^{*}(s))}p(z|s)$.
Since the bellman update converges from any bounded $Q^{(0)}$, taking $k \to \infty$, we obtain
\begin{align*}
Q^*_{/{\bf z}}(s,a)= Q^*_{z}(s,a) + M^*_z(s,a) 
\end{align*}
where $M^*_z(s,a)$ is the fixed point of the recursion 
\begin{align*}
M^{(k+1)}(s,a) = \mathbb{E}_{s' \sim p(s'|s,a)}[-\log(p(z|s'))+ \mathbb{E}_{a' \sim \pi_{/{\bf z}}(\cdot|s')}[M^{(k)}(s',a')]
\end{align*}
\end{proof}
\end{thm}

\section{Experimental Details}
For the Multigoal, Hopper, and Walker environments, we represent separate agents by concatenating a one hot encoding of $z$ to the state thereby conditioning the policy and $Q$ networks.  Policy and $Q$ functions are represented by neural networks with two layers of $128$ units and ReLu activations in the Multigoal environment and two layers with $300$ units and ReLu activation in the Hopper and Walker environments. For all three of these domains, we use an entropy temperature coefficient of $.3$ and no reward scaling.

We found that we could not represent truly diverse Multidirection Ant policies with the same architecture as the Multigoal, Hopper and Walker experiments.  All policies would eventually converge to a particular direction and run quickly in that direction (much like SAC as discussed in the main paper). In order to represent the diversity of running in all different directions, we use completely separate policy and $Q$ function networks for each agent with two hidden layers of $128$ units and ReLu activations.  For efficiency, each agent interacts with the environment and trains within a separate thread. There is an additional thread training the discriminator on batches sampled from the replay buffer of each agent.  After the discriminator is trained, each agent syncs a local copy of the discriminator to use for its own training. In this domain, we use an entropy temperature coefficient of $1$ and reward scaling of $5$.

Across all experiments, policy networks use a Gaussian output layer, the discriminators are represented by neural networks with two layers of $256$ units and ReLu activations and networks are trained on single batch of size $256$ per environement step.

\end{document}